\documentclass{article}

\usepackage{arxiv}

\usepackage[utf8]{inputenc} 

\usepackage{amsfonts}       
\usepackage{nicefrac}       
\usepackage{microtype}      
\usepackage{lipsum}

\usepackage{color}
\usepackage{listings}
\usepackage{booktabs}
\usepackage{array}
\usepackage{graphicx}
\usepackage{longtable}
\usepackage{subfigure}

\usepackage{cite}
\usepackage{url}
\usepackage{fancyhdr}

\usepackage{soul}

\usepackage{float}
\usepackage{listings}
\usepackage{mdwmath}
\usepackage{mdwtab}
\usepackage{multirow}
\usepackage{multicol}
\usepackage{rotating}
\usepackage{setspace}
\usepackage[utf8]{inputenc}
\usepackage{lineno}
\usepackage{listings}

\usepackage{mdwmath}
\usepackage{mdwtab}
\usepackage{multirow}
\usepackage{multicol}
\usepackage{array}
\usepackage{booktabs}

\usepackage{enumitem}
\usepackage{xspace}
\usepackage[export]{adjustbox}
\usepackage{graphicx}
\usepackage{color,soul}
\usepackage{rotating}
\usepackage{setspace}
\usepackage{amsmath} 
\usepackage{amssymb}
\usepackage{float}
\usepackage{xcolor}

\usepackage{hyperref}
\usepackage{url}

\title{A Reference Architecture for Designing Foundation Model based Systems}

\author{Qinghua Lu\thanks{Contact email: qinghua.lu@data61.csiro.au}, Liming Zhu, Xiwei Xu, Zhenchang Xing, Jon Whittle\\
Data61, CSIRO, Australia
}

\begin{document}

\maketitle

\begin{abstract}
The release of ChatGPT, Gemini, and other large language model  has drawn huge interests on foundations models. There is a broad consensus that foundations models will be the fundamental building blocks for future AI systems. However, there is a lack of systematic guidance on the architecture design. Particularly, the the rapidly growing capabilities of foundations models can eventually absorb other components of AI systems, posing challenges of moving boundary and interface evolution in architecture design. Furthermore, incorporating foundations models into AI systems raises significant concerns about responsible and safe AI due to their opaque nature and rapidly advancing intelligence. To address these challenges, the paper first presents an architecture evolution of AI systems in the era of foundation models, transitioning from "foundation-model-as-a-connector" to "foundation-model-as-a-monolithic architecture". The paper then identifies key design decisions and proposes a pattern-oriented reference architecture for designing responsible foundation-model-based systems. The patterns can enable the potential of foundation models while ensuring associated risks.

\end{abstract}

\textbf{Key terms - } Responsible AI, ethical AI, AI safety, architecture, pattern, foundation model, large language model, LLM, ChatGPT, AGI, GenAI.

\section{Introduction}


{T}{he} release of ChatGPT, Bard, and other large language model (LLM)-based applications has drawn huge attention on foundation models (FMs) worldwide. FMs are massive AI models that are pre-trained on vast amounts of broad data and can be adapted to perform a wide variety of tasks~\cite{bommasani2021opportunities}. With numerous projects already underway to explore their potential, it is widely predicted that FMs will serve as the fundamental building blocks for most future AI and artificial generative intelligence (AGI) systems. 

Many reusable solutions have been proposed to tackle various challenges in designing FM-based systems. However, there is a lack of systematic guidance on the architecture design of FM-based systems. The impact of integrating FMs into software architecture are not fully studied yet. Additionally, the FM's growing capabilities can eventually absorb the other components of AI systems, introducing the moving boundary and interface evolution challenges in architecture design.

\begin{figure*}
\centering
\includegraphics[width=\textwidth]{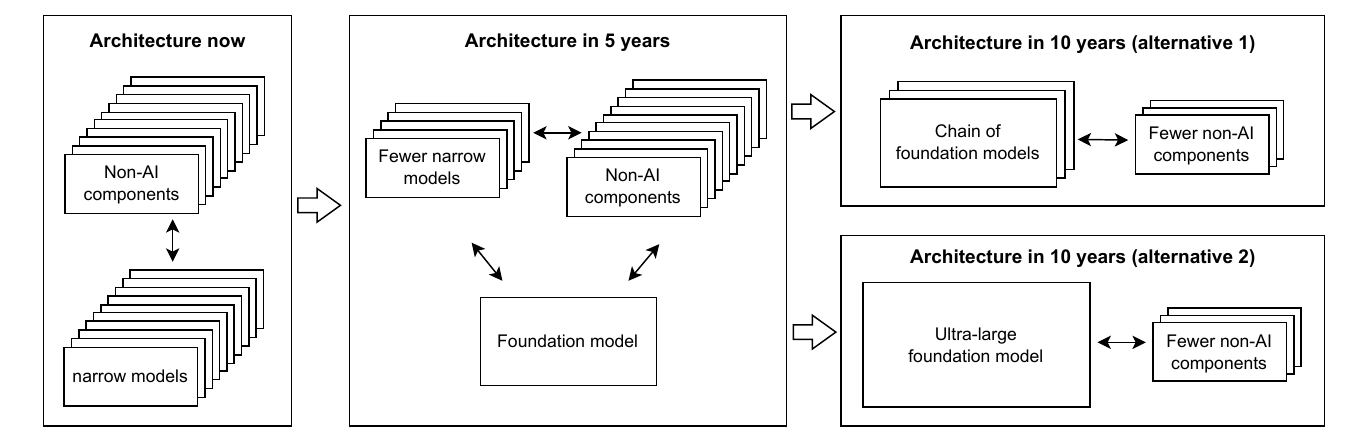}
\caption{Architecture evolution: from "foundation-model-as-a-connector" to "foundation-model-as-a-monolithic-architecture".} \label{fig:architecture}
\vspace{-2ex}
\end{figure*}

On the other hand, there are unique challenges on building responsible and safe AI into the architecture of FM-based systems~\cite{RAIbook}. First, accountability becomes more complex due to the involvement of multiple stakeholders. The accountability for decisions made by FM-based systems may be shared among the system owner, the FM provider, and various providers of external knowledge bases and tools (such as ChatGPT plugins~\footnote{https://openai.com/blog/chatgpt-plugins}). 
Enabling accountability necessitates the underlying supporting mechanisms for traceability. It is essential to record the inputs and outputs of FMs, non-FM components (e.g., guardrails), and external tools, services, and systems.
Second, trustworthiness is a substantial obstacle in designing the FM-based systems, e.g., whether the user prompts or the FM reasoning processes and outputs align with human goals and fulfill trustworthiness criteria. 
Third, the potential AI/AGI misuse poses a considerable challenge, which requires continuous risk assessment to ensure the instructions for the FM-based systems set by humans are responsible and safe.  


There is an urgent need for concrete system-level guidance to design responsible FM-based systems. In this paper, we first discuss the potential architecture evolution of AI systems in the era of FMs and highlight the key quality attributes necessary for the design of responsible and safe FM-based systems. We then identify the major decision making points when designing FM-based systems. Finally, we propose a pattern-oriented reference architecture, which provides a responsible-AI-by-design architectural template for designing FM-based systems and considers the evolution of architecture to ensure adaptability over time.

\section{Architecture evolution of AI systems}
FMs are designed to provide a wide range of comprehensive capabilities that can be applied to various tasks, rather than being limited to specific functionalities~\cite{bommasani2021opportunities}. One key challenge that the architecture design of FM-based systems faces is that FMs could eventually absorb functionalities and tools that were originally external components. While these components may exist for a while, they can become short-lived and eventually get integrated into the FM, resulting in a single, monolithic blob at the center of the architecture. 
As illustrated in Fig.\ref{fig:architecture}, the architecture evolution of AI systems can be divided into three stages:

\begin{itemize}
    \item \textbf{\textit{Architecture now}: many narrow AI models + many non-AI components}. The current architecture of AI systems usually comprises narrow models focusing on specific tasks and non-AI components. These narrow models and non-AI components co-exist within the architecture of the AI systems and interact with each other to enable the systems to function properly. The narrow models are responsible for processing data and making inference, while the non-AI components are responsible for tasks such as user interface, data storage, interaction with other systems. 
    \item \textbf{\textit{Architecture in 5 years} - FM-as-a-connector: 1 FM + fewer narrow models + many non-AI components}. In this architecture, the FM acts as a connector between external components, i.e., narrow models or non-AI components. The FM can provide four types of connector services~\cite{lu2023framework}: 
    \begin{itemize}
        \item \textbf{FM-as-a-communication-connector}: enabling the transfers of data between software components, e.g., extracting the task description from the user prompt and transferring to other components for further processing.
        \item \textbf{FM-as-a-coordination-connector}: planing a workflow and coordinating task execution through various software components.
        \item \textbf{FM-as-a-conversion-connector}: functioning as an interface adapter for software components that use different data formats to communicate with each other, e.g. parse the task into machine-readable template for executing by an AI model. 
        \item \textbf{FM-as-a-facilitation-connector}: facilitating the interactions between components, e.g., creating logs or deciding the invocation of local models.
    \end{itemize}
    In such architecture, FMs still need to interface with narrow AI models and non-AI components to tackle complex tasks, such as HuggingGPT~\cite{shen2023hugginggpt}. However, as the capabilities of FMs continue to expand rapidly, it is expected that many of those components will be eventually absorbed into the FMs and ultimately disappear. 

    \item \textbf{\textit{Architecture in 10 years}}:
    \begin{itemize}
        \item \textbf{Alternative 1: chain of FMs + fewer non-AI components}. There is a chance that most of the software components could be absorbed into the FMs. Thus, one alternative of the architecture in 10 years is a modularised architecture, such as Socratic Models~\cite{zeng2022socratic}. This architecture relies on a few FMs that are chained together and a limited number of non-AI components to perform tasks (e.g., through language-based interactions) without requiring additional training or fine-tuning. The inference for a task-specific output is jointly performed by multimodal interactions between the independent FMs, such as text-to-text FMs, text-to-visual FMs and text-to-audio FMs. Those FMs can be connected via APIs with external non-AI components that offer additional functionalities, such as robotic systems or web search engines. By multimodal interaction between independent FMs, the architecture can effectively leverage the capabilities of different FMs and external non-AI components. 
        In this architecture, prompt engineering is important for guiding the FM to produce high-quality responses. Various prompt patterns can be applied, including few-shot prompting, self-consistency, chain-of-thought, retrieval augmented generation, etc~\cite{lu2023framework}. 
        
        \item \textbf{Alternative 2 - 1 ultra-large FM + fewer non-AI components}. Another potential type of future architecture is a monolithic architecture, which only contains a single big FM capable of performing a variety of tasks by incorporating different types of sensor data for cross-training. An example of this type of architecture is PaLM-E~\cite{driess2023palm}, which is used for performing language, visual-language, and reasoning tasks. In this type of architecture, no external components are required, including prompt components. In this architecture, the non-AI components may include context engineering components (such as multimodal context injection), prompt engineering components (such as prompt optimiser), and responsible AI components (such as continuous risk assessment). 
    \end{itemize}
\end{itemize}

As FMs continuously and rapidly evolve with growing capabilities, many of the existing software components will be likely to become obsolete since their functions will be provided by new versions of FMs. For example, Tesla AI is working on an end-to-end model which learns all steps from the initial input phase and the output result phase~\footnote{\url{https://twitter.com/Tesla_AI/status/1730761835694153790}}. This means that context data goes in and driving decisions come out, without a single line of code implemented in the process of autonomous driving. Thus, adaptability and modifiability are the two key concerned software quality attributes. Adaptability refers to a software system's ability to adapt to run-time changes in its environment without requiring external intervention~\cite{ISO}, such as changes in the data being processed. Modifiability is the ease with which a software system can be changed at static-time~\cite{ISO}, such as adding new features, fixing bugs, or changing the underlying infrastructure. Both adaptability and modifiability are important qualities attributes for an evolving architecture, as they can significantly impact the long-term maintainability of a system. The patterns of conventional software systems could be applied to manage the issues of moving boundary and interface evolution in the FM-based systems.

 

\section{Architectural Design decisions}\label{sec:design}
There are some major architectural design decisions that developers need to consider when building FM-based systems.
\subsection{Design decision 1: Different design options for using FMs} 
When designing the architecture, one of the most important decisions is choosing which type of FM to use. There are three types of FMs in terms of sourcing: 
\begin{itemize}
    \item \textbf{FM type 1: sovereign (self-developed) FMs.} Sovereign FMs can have complete control over data and model training and ensure responsible AI. Also, some organisations may possess unique internal data and training a sovereign FM in-house from scratch can become their unique competitive advantage. However, it requires high investment in cost and resources, including data, computational, and human resources.
    \item \textbf{FM type 2: fully externally sourced FMs.} As this type of FMs are pre-trained by external organisations on massive data with numerous computational resources, using these models can be significantly cheaper than training one from scratch. To improve the accuracy of the externally sourced FMs, in-context learning is often employed by integrating context information such as input-output examples through prompts. However, there are still accuracy and responsible AI related issues with the outputs due to token limits in the context window. For example, the latest version of GPT-4 has a maximum limit of 128k tokens, which is equivalent to approximately 300 pages of text~\footnote{\url{https://platform.openai.com/docs/models/continuous-model-upgrades}}. This can restrict the FM's ability to understand complex context information. Alternatively, another design option is retrieval augmented generation (RAG), which uses a vector database for storing the internal data as vector embeddings, which can be used to perform similarity searches and enable the retrieval of internal data that are related to specific prompts.
    \item \textbf{FM type 3: externally sourced but self-customised FMs.} Fine-tuning with labeled target data is a common method to customise FMs. Full fine-tuning retrains all the parameters, which is less feasible and extremely expensive (such as GPT-3 with 175B parameters). Another way is to employ parameter-efficient fine-tuning techniques, which focuses on retraining a smaller subset of parameters. However, fine-tuning is highly coupled with the open source FMs, which are not portable. Also, responsible AI issues may still exist due to the lack of control over the external FM training process.
    
\end{itemize}


    
    

\subsection{Design decision 2: Chain of FMs vs. an ultra-large FM} 
When considering FMs developed by external organisations, one important decision is whether to use a chain of models (such as Socratic Models~\cite{zeng2022socratic}) or an ultra-large FM (such as PaLM-E~\cite{driess2023palm}). The chain of FMs generates joint predictions and offers an modularised architecture that allows for easy switching to other FMs with specific capabilities, e.g., switching to a more powerful visual language model to improve performance. 
This option may improve maintainability, but it may come with an additional cost to understand the capabilities and limitations of different foundations models.
On the other hand, using ultra-large FM may achieve better performance via cross-training on numerous multi-modal data~\cite{driess2023palm}. However, this option may come at the cost of reduced maintainability. There may be a risk for vendor lock-in, as there may be few providers in the market with similar capabilities. 
It is challenging to determine which option is better, and experimentation is necessary to evaluate each option's effectiveness for a specific context.

\subsection{Design decision 3: Responsibilities of external components} 
Responsibility is a concept in a software context that comes from object-oriented design. A responsibility can be an action,
a piece of knowledge to be maintained, or a decision to be carried out by a software component. 

FMs can gradually absorb external components by taking on their responsibilities over time. This can create a moving boundary issue where the responsibility of a software component shifts from an external component to the FM. To address this issue, one key design decision is to determine the responsibilities of software components. The responsibility can be split into a bunch of smaller responsibilities that are placed in distinct components. Changes can be isolated to specific components, making it easier to manage the external component that could be absorbed by the FM over time. As FMs are built around capabilities~\cite{bommasani2021opportunities}, it may be worth breaking down a large component along capability lines. This allows the developers to choose which FM's capabilities to use, e.g., use a good enough one or an emerging new one.

Breaking down responsibilities into smaller components can improve adaptability and modifiability, ensuring long-term maintainability. However, it can also introduce additional communication overhead between smaller components, as each component may need to interact with other components to accomplish tasks. Additionally, it can make the system more complex and difficult to understand how the components work together, potentially reducing maintainability. 

\subsection{Design decision 4: Automatic response vs. verifier} 
When designing FM-based systems, an important consideration is how to ensure the systems' responses are accurate and responsible. One option is to rely solely on the FM to generate responses to user queries. While this option can be efficient and cost-effective, it may result in inaccurate or irresponsible responses that can affect user trust or cause harm.

To address this issue, before responding to the user, a verifier can be adopted to verify whether the FM's reasoning processes and outputs meet the specified requirements such as functional and trustworthiness requirements. The verifier could be a human verifier or an AI verifier (FM-based or non-FM-based)~\cite{lu2023framework}. This is particularly useful when designing FM-based agents. The verification process and results can be stored in an agent's long memory to help its continuous learning and improvement. However, verifying the reasoning processes and outputs of FMs directly can be challenging. It is necessary to take a conversational step-by-step approach to conduct verification. The choices between automatic response and verifier depends on the system's priorities and the consequences of inaccurate or irresponsible responses.
For systems where accuracy and trustworthiness are critical, a verifier approach may be the better option, even if it comes at a higher cost. However, for systems where efficiency and cost-effectiveness are more important, an automatic response approach may be more suitable, with periodic checks.

\begin{figure*}
\centering
\includegraphics[width=\textwidth]{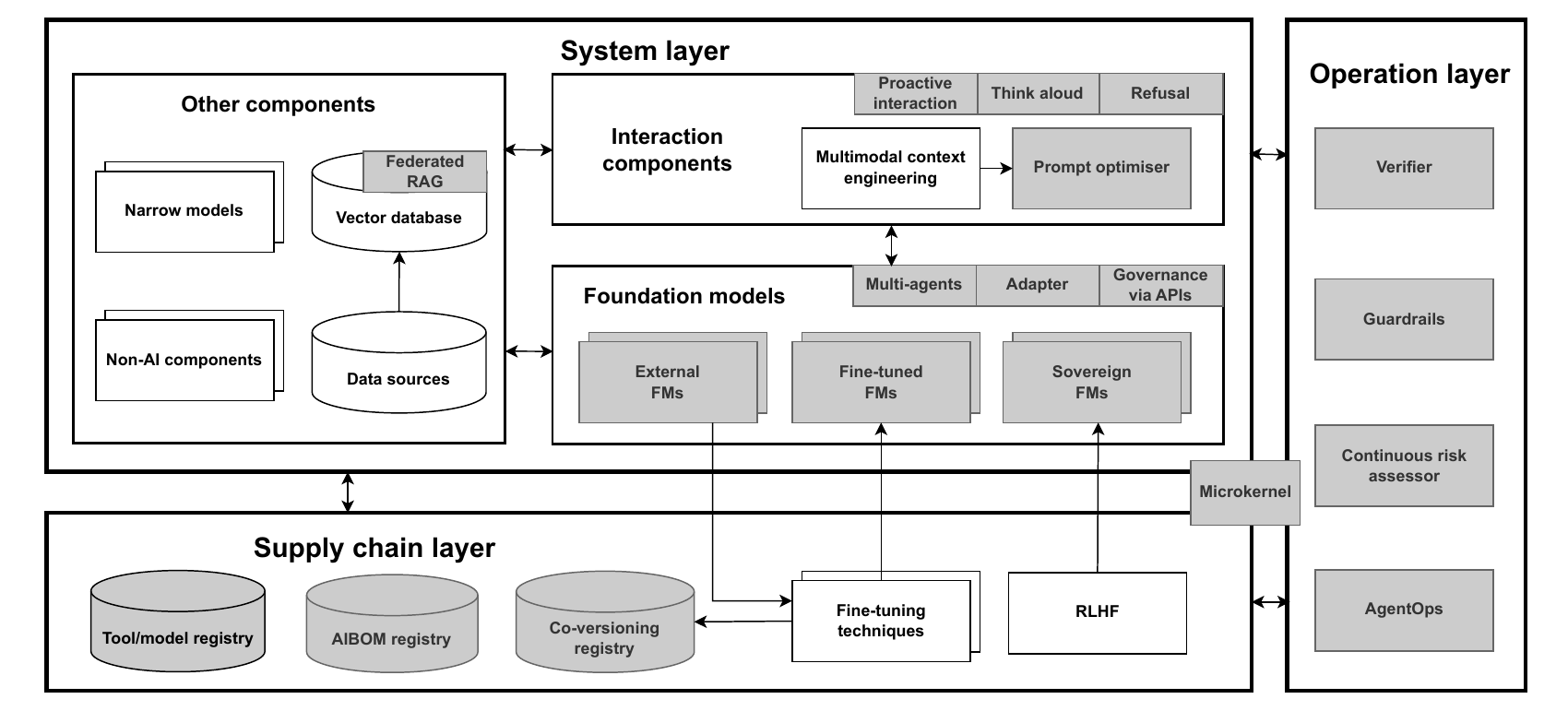}
\caption{A pattern-oriented reference architecture for designing responsible FM-based systems.} \label{fig:FMarchitecture}
\vspace{-2ex}
\end{figure*}

\subsection{Design decision 5: Passive interaction vs. proactive interaction} 
Interaction in FM-based systems involves context engineering and prompt engineering. Context engineering aims to gather and structure the context in which the FM-based systems operate to understand the users' goals or intentions~\cite{xia2023towards}, while prompt engineering creates prompts that act as guiding instructions, enabling the FM-based systems to effectively complete human's tasks and goals. 

There are two distinct interaction patterns: passive interaction and proactive interaction. Passive interaction interprets the user's intentions as described through text prompts submitted via the dialogue interface. In contrast, proactive interaction anticipates the user's intentions and makes proactive suggestions by understanding multimodal context data, such as platform UI elements, screen recording~\cite{zhao2023seehow}, mouse clicks~\footnote{\url{https://github.com/ddupont808/GPT-4V-Act}}, typing, eye tracking, gestures~\cite{zeng2023gesturegpt}, document annotations and notes. Compared to passive interaction, proactive interaction introduces a deeper level of autonomy.

\subsection{Design decision 6: Single agent vs. multi-agents} 
The tasks that need to be performed by an FM can vary in complexity and scope. For simple tasks or goals, such as answering frequently asked questions, a single agent may be sufficient. However, for more complex tasks, it may be necessary to use multi-agents to ensure the performance. For example, one agent can be responsible for generating the plan which may include multiple tasks, while other agents can be designed for completing the tasks by leveraging different tools. However, using multi-agents can also introduce communication costs between agents and increase the level of design complexity.

\subsection{Design decision 7: Think aloud vs. think silently} 
There are two options to consider when it comes to the explaining the decision-making and execution process: think aloud and think silently. The think aloud design pattern can be used disclose the intermediate steps, such as the prompt optimisation process, the reasoning process, the task completion status, etc. This design can help build human trust in the system, but it may sacrifice data privacy. For example, some system providers may view the prompt optimisation as business sensitive data and intelligence property. In such case, they may need to carefully consider which parts of the intermediate process they are willing to share with the users.

\section{Reference architecture}\label{sec:architecture}
Fig.~\ref{fig:FMarchitecture} illustrates a pattern-oriented reference architecture for designing responsible and adaptable FM-based systems. The architecture comprises three layers: the system layer, which includes the components of the deployed AI system, the operation layer, which provides responsible AI tooling functions to the AI system, and the supply chain layer, which generates the software components that compose the AI system. The grey-coloured boxes and cylinders are the components where the design patterns are applied. 
An empirically-grounded design methodology has been adopted for designing the reference architecture~\cite{galster2011empirically}. The type of our reference architecture is an industry-crosscutting, classical, facilitation reference architecture. Our design strategy is a combination of research-driven and practice-driven, as the design of this reference architecture is founded mainly on the findings of literature review~\cite{lu2022responsible} and our project experience\footnote{\url{https://research.csiro.au/ai4m/operationalising-responsible-ai/}}.

\subsection{System layer}
The system layer comprises the components of the deployed FM-based systems.
Interaction components comprises two sub-components: \textbf{multimodal context engineering} and \textbf{prompt optimiser}. Multimodal context engineering is designed to collect and structure the context in which the system operates to understand the user's goals or tasks.
Instead of analysing user's goals or tasks described through text prompts sent by the user via the dialogue interface, the system can proactively anticipate the user's goals by analysing multimodal context information, including screen recording~\cite{zhao2023seehow}, mouse clicks, typing, eye tracking, gestures~\cite{zeng2023gesturegpt}, document annotations and notes. A user prompt may lack of relevant context or contain unintentional injection attacks, etc.
\textbf{Prompt optimiser} can construct refined prompts according to relevant standards, specifications, etc. A prompt template is often used as a factory for creating prompt instances. The template provides a structured way to standardise the queries, which can improve the response accuracy and interoperability with external systems.\textbf{Think aloud} is a pattern for explainability, which can describe the system's capabilities, limitations, the rationale behind its intermediate or final outputs, and ethical or legal implications. This design pattern can help build human trust in the system, but it may sacrifice data privacy. \textbf{Prompt refusal} filters inappropriate or harmful tasks, e.g., refusing to generate responses that contain violence promotion or hate speech.

There are several patterns for using FMs, including the \textbf{external FMs}, \textbf{fine-tuned FMs}, \textbf{sovereign FMs}, \textbf{chain of FMs} and \textbf{ultra-large FMs}. 
Due to the rapidly growing capabilities of FMs, there are issues with the moving boundary and interface evolution in architecture design. Most of the components in the system layer, operation layer, and supply chain layer will eventually be absorbed by FMs. In some cases, absorption is not done component by component, but rather by splitting a component in the middle. To address these issues, two classic patterns can be applied to ensure adaptability and modifiability: \textbf{microkernel} pattern~\cite{bachmann2007modifiability} and \textbf{adapter} pattern (also known as \textbf{wrapper}). \textbf{Microkernel} pattern can help place smaller responsibilities in distinct component so that changes can be isolated to specific components, which could be absorbed by the FM overtime.
When a component is absorbed by an FM, the component's original connector used to communicate with other components may need to be converted into a certain format of interface (such as a text interface for LLMs) through the use of an \textbf{adapter} pattern.

Autonomous agents can take a proactive, autonomous role to pursue users' goals. These agents derive their autonomy from the capabilities of FMs, enabling them to break down the given goal into a set of manageable tasks and orchestrate task execution to fulfill the goal. The agents can be categorised into two types of roles: agent-as-a-coordinator and agent-as-a-worker. Agents in the coordinator role primarily formulate high-level strategies and orchestrate the execution of tasks by delegating task execution responsibilities to \textbf{multi-agents} in the worker role. These agents in the worker role need to generate strategies and execute specific tasks in line with those strategies. To complete these tasks, agents in the worker role may need to cooperate or compete with other agents, or call external tools or non-agent AI/non-AI systems.

To prevent harmful dual-use of AI systems, developers should impose restrictions on their usage and prevent users from getting round of restrictions through unauthorised reverse engineering or modification of the system design. One way to do this is by implementing \textbf{governance via APIs} pattern, which involves providing AI services on cloud platforms and managing interactions through API controls (such as GPT4), rather than allowing AI systems to run locally with unrestricted access.

To improving the response accuracy of FMs, retrieval augmented generation (RAG) can be adopted by using a ~\textbf{vector database} (such as Pinecone~\footnote{https://www.pinecone.io}) for storing the domain data from various data sources as vector embeddings. These embeddings can be used to perform similarity searches and enable the retrieval of data that are related to specific tasks. When the tasks involves cross-organisational data analytics, \textbf{federated RAG} can be applied by adapting federated learning. Each organisation deploys an FM and has its own RAG in which the data is confidential to other organisations. The data in the local RAGs can be aggregated and further processed by a central FM.


\subsection{Operation layer}
The operation layer includes components responsible for monitoring and managing the responsible AI related qualities in deployed FM-based systems.
A \textbf{verifier}, whether human or AI (FMs or non-FM models)~\footnote{https://openai.com/blog/our-approach-to-alignment-research}, can be introduced to check whether the final or intermediate outputs meet the specified requirements such as topic requirements or trustworthiness requirements. 

\textbf{Guardrails} can be implemented at three stages: 1) Preprocessing: After receiving user prompts, guardrails can be enforced after verifying whether the prompts can comply with responsible AI requirements through the verifier, e.g., rejecting prompts outside the pre-configured scope or removing personal identifiable information (PII) before sending them to the FMs. 2) Intermediate process: During task execution, guardrails monitor the FM's reasoning process, data extracted through RAG, external tools invoked, or the output at each intermediate step. If they do not meet the users' specific requirements, including trustworthiness requirements, specific guardrails should be triggered. 3) Postprocessing: When the FMs return results, guardrails are necessary to ensure that the outputs meet the requirements including responsible AI requirements, structure requirements, etc. 

The design of guardrails should be dynamic, which means they can learn during operation time. Various quality attributes should be considered when designing guardrails: \textit{generalisability}, \textit{customisability}, \textit{performance}, \textit{interpretability}, and \textit{portability}. Additionally, guardrails should be capable of processing multimodal data, such as text, audio, video, etc.

A \textbf{continuous risk assessor} continuously monitors and assesses AI risk metrics~\footnote{https://oecd.ai/en/catalogue/metrics} to prevent the misuse of the agent and to ensure the trustworthiness of the agent. For example, when a user submits a prompt through the dialogue interface, the continuous risk assessor can assess the potential risks of the intended goals and may modify or reject the prompt based on the risk assessment results. 
\textbf{AgentOps} can be viewed as a large logging repository which records the runtime data and shared with relevant stakeholders to enable traceability. The recorded data includes the input, output, and intermediate data for each layer or component within the architecture, such as prompt version, the input and output of the FMs, guardrails, or other components. 
All these data need to be kept as evidence with the timestamp and location data. Design decisions need to be made on what data should be recorded and where the data should be stored (e.g., using a blockchain-based immutable log or a cloud-based data storage). 

\subsection{Supply chain layer}

The supply chain layer includes all components involved in developing and procuring both AI components (including FMs) and non-AI components. 
FM providers can use reinforcement learning from human feedback (RLHF) to fine-tune the FM's behaviour and produce more accurate and responsible responses. RLHF allows humans to provide feedback on the quality of the responses and uses this feedback to adjust the model's parameters. The FM is then trained to maximise the reward it receives from human feedback, which can improve its accuracy and responsible AI related qualities over time.
Parameter-efficient fine-tuning techniques, e.g., LoRA~\footnote{https://github.com/microsoft/LoRA}, can reduce the number of training parameters by 10,000 times and decreasing GPU usage by threefold. The external tools or models information, such as API or model description, can be maintained in the \textbf{tool/model registry}.
All components procured from third parties can be associated with a bill of materials (BOM) that records their supply chain details, which can include responsible AI (RAI) metrics or verifiable RAI credentials. This procurement information can be maintained in an \textbf{AIBOM registry}. FMs can refuse to call the third party components or tools/models that fail to provide registered AIBOM information.
To ensure auditability, the \textbf{co-versioning registry} pattern can be applied to co-version the AI artifacts, such as external/sovereign FMs, fine-tuned FMs, and training/testing datasets.

\section{Evaluation}
In this section, we evaluate the completeness and utility of the proposed reference architecture by mapping it to the architecture of a real-world FM-based system, responsible AI (RAI) chatbot. This chatbot enables scientists to understand and assess potential AI risks in their AI projects\footnote{https://research.csiro.au/ai4m/operationalising-responsible-ai/}. The current version is built on GPT-4, an ~\textbf{external FM} provided by OpenAI. Users interact with the chatbot by describing their AI systems and asking questions through a web interface. Questions not relevant to RAI are automatically rejected through \textbf{prompt refusal}. 

To improve accuracy in responding to user queries, the RAI chatbot employs \textbf{RAG} by integrating LlamaIndex~\footnote{https://www.llamaindex.ai}, which connects GPT-4 with local \textbf{data sources}. The local data sources include a manually labelled AI incident database, the Responsible AI Question Bank~\cite{lee2023qb4aira}, and the Responsible AI Pattern Catalogue~\cite{lu2022responsible}. 

The chatbot incorporates the \textbf{think aloud} pattern, making the intermediate processes transparent to users. \textbf{Multi-agents} are used by creating multiple instances to perform different tasks. For example, one agent is responsible for question-answering interactions, while another summarises the documents shared by users. All conversation histories are captured by a \textbf{black box recorder}. The \textbf{verifier} pattern is implemented, allowing a human expert to review and edit the answers generated by GPT-4. The project team is currently building an AIBOM registry for the RAI chatbot and discussing the potential of fine-tuning an FM to further increase response accuracy.

It can be concluded that our reference architecture is complete and usable as the architecture of RAI chatbot is successfully mapped to the proposed reference architecture. We have observed that the fundamental layers in an FM-based system architecture include the system layer, operation layer and supply chain layer. The key components in the system layer include FMs, interaction components, data sources. 

\section{Conclusion}
This paper presents pattern-oriented responsible-AI-by-design reference architecture to address the challenges of responsible AI and architecture evolution in FM-based systems. We first discuss the architecture evolution and identify two important software qualities for building FM-based systems: adaptability and modifiability. Then, we summarise seven key design decisions in architecture design and discuss the trade-offs between responsible AI related software qualities. Finally, we present a pattern-oriented reference architecture to provide a concrete guidance for developers to design responsible and adaptable FM-based systems. In the future, we will build a pattern catalogue for designing FM-based systems/agents.




\end{document}